\documentclass[11pt]{article}
\usepackage{coling2020}
\usepackage{times}

\usepackage{hyperref}
\usepackage{url}
\usepackage{graphicx}
\usepackage{subfigure}
\usepackage{booktabs}
\usepackage{amsmath}
\usepackage{multicol}
\usepackage{multirow}
\usepackage{mathtools}
\usepackage{color}

\DeclareMathOperator*{\onedcnn}{1D-CNN}

\DeclareMathOperator*{\pool}{Pool}
\DeclareMathOperator*{\mlp}{MLP}
\DeclareMathOperator*{\mysigmoid}{sigmoid}
\DeclareMathOperator*{\layernorm}{LayerNorm}

\usepackage{amsmath,amsfonts,bm}

\def\eqref#1{equation~\ref{#1}}
\def\1{\bm{1}}

\def\vb{{\bm{b}}}
\def\vc{{\bm{c}}}

\def\vh{{\bm{h}}}

\def\vp{{\bm{p}}}

\def\vs{{\bm{s}}}

\def\vu{{\bm{u}}}
\def\vv{{\bm{v}}}
\def\vw{{\bm{w}}}
\def\vx{{\bm{x}}}

\def\mA{{\bm{A}}}

\def\mH{{\bm{H}}}

\def\mR{{\bm{R}}}

\def\mU{{\bm{U}}}
\def\mV{{\bm{V}}}
\def\mW{{\bm{W}}}
\def\mX{{\bm{X}}}

\DeclareMathAlphabet{\mathsfit}{\encodingdefault}{\sfdefault}{m}{sl}
\SetMathAlphabet{\mathsfit}{bold}{\encodingdefault}{\sfdefault}{bx}{n}

\def\gD{{\mathcal{D}}}

\def\gL{{\mathcal{L}}}

\def\sR{{\mathbb{R}}}

\newcommand{\R}{\mathbb{R}}

\newcommand{\softmax}{\mathrm{softmax}}

\colingfinalcopy % Uncomment this line for the final submission

\title{
Improving Long-Tail Relation Extraction \\with Collaborating Relation-Augmented Attention
}

\author{
Yang Li$^1$, Tao Shen$^1$\thanks{~~Corresponding author.}, Guodong Long$^1$, Jing Jiang$^1$, Tianyi Zhou$^2$ \and Chengqi Zhang$^1$ \\
$^1$Australian AI Institute, School of Computer Science, FEIT, University of Technology Sydney \\
$^2$University of Washington, Seattle\\
\texttt{\{yang.li-17,tao.shen\}@student.uts.edu.au}, \texttt{tianyizh@uw.edu} \\
\texttt{\{guodong.long,jing.jiang,chengqi.zhang\}@uts.edu.au}
  }

\date{}

\begin{document}
\maketitle
\begin{abstract}
Wrong labeling problem and long-tail relations are two main challenges caused by distant supervision in relation extraction. Recent works alleviate the wrong labeling by selective attention via multi-instance learning, but cannot well handle long-tail relations even if hierarchies of the relations are introduced to share knowledge. In this work, we propose a novel neural network, Collaborating Relation-augmented Attention (CoRA), to handle both the wrong labeling and long-tail relations. Particularly, we first propose relation-augmented attention network as base model. It operates on sentence bag with a sentence-to-relation attention to minimize the effect of wrong labeling. Then, facilitated by the proposed base model, we introduce collaborating relation features shared among relations in the hierarchies to promote the relation-augmenting process and balance the training data for long-tail relations. Besides the main training objective to predict the relation of a sentence bag, an auxiliary objective is utilized to guide the relation-augmenting process for a more accurate bag-level representation. In the experiments on the popular benchmark dataset NYT, the proposed CoRA improves the prior state-of-the-art performance by a large margin in terms of Precision@N, AUC and Hits@K. Further analyses verify its superior capability in handling long-tail relations in contrast to the competitors. 
\end{abstract}

\section{Introduction} \label{sec:intro}
Relation extraction, as a fundamental task in natural language processing (NLP), aims to discriminate the relation between two given entities in a plain text. As recent data-driven algorithms, e.g., deep neural networks \cite{han2018hierarchical,du2018multi,ye2019distant}, have shown their capabilities in tackling NLP tasks, labor-intensive annotation and scarce training data become the major obstacles of achieving promising performance on relation extraction. 

Instead of costly manual labeling, distant supervision method \cite{mintz2009distant} is proposed to auto-label the training data for relation extraction. It labels a sentence containing a pair of entities with the relation between them in a knowledge graph, e.g. Freebase \cite{bollacker2008freebase}. A strong assumption behind this is that a sentence containing two entities only expresses the relation existing in the knowledge graph, but this assumption will not always hold \cite{lin2016neural}. 
Hence, it leads to two main problems regrading the training data. 
First, when the assumption is invalid, \textit{wrong labeling problem} appears and degrades algorithms by introducing noisy supervision signals. 
This problem has been well-studied by recent works \cite{lin2016neural,han2018hierarchical,ji2017distant,li2020self} operating at bag level for multi-instance learning \cite{hoffmann2011knowledge}, where the ``bag'' denotes a set of sentences with the same entity pair. Most of them use selective attention to avoid wrongly-labeled sentences. %  in a bag. 
%  with multi-instance learning. 

\begin{figure}[t]
	\centering
	\subfigure
	{
		\includegraphics[width=0.263\textwidth]{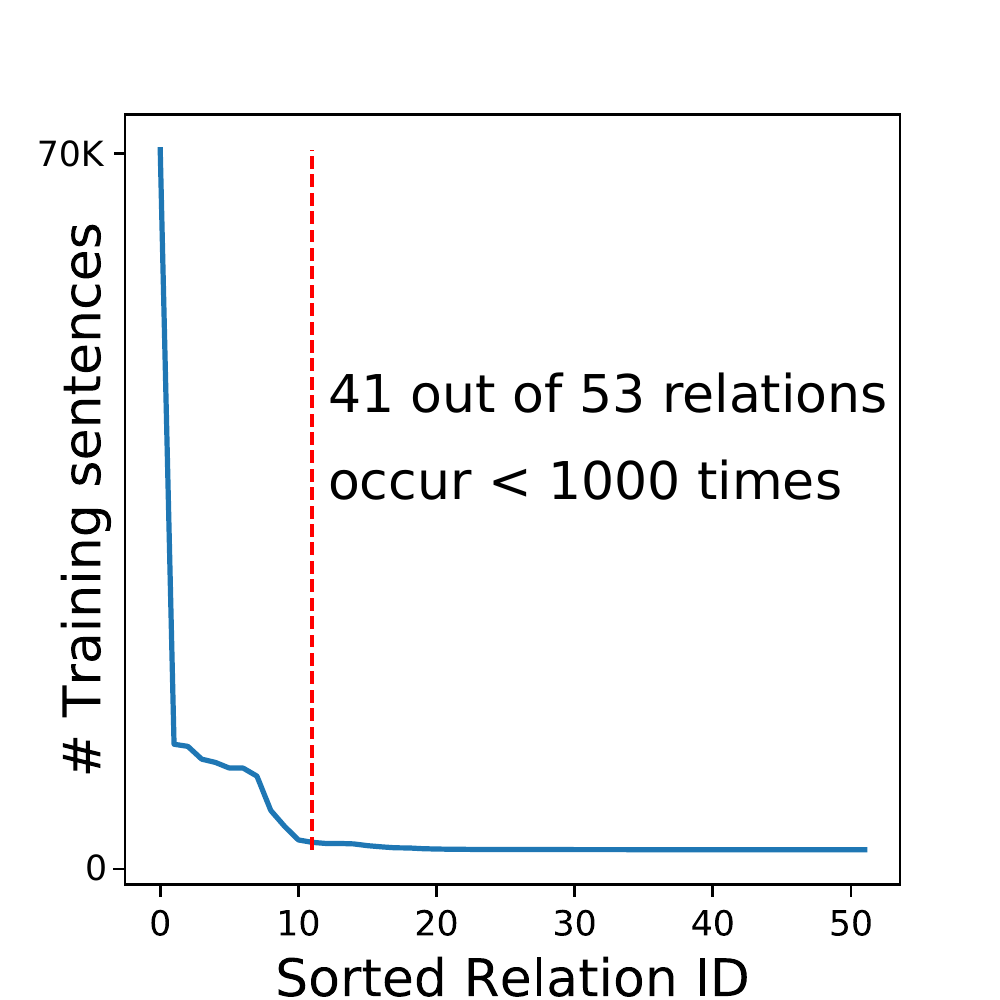}
    }
    \subfigure
    {
        \includegraphics[width=0.420\textwidth]{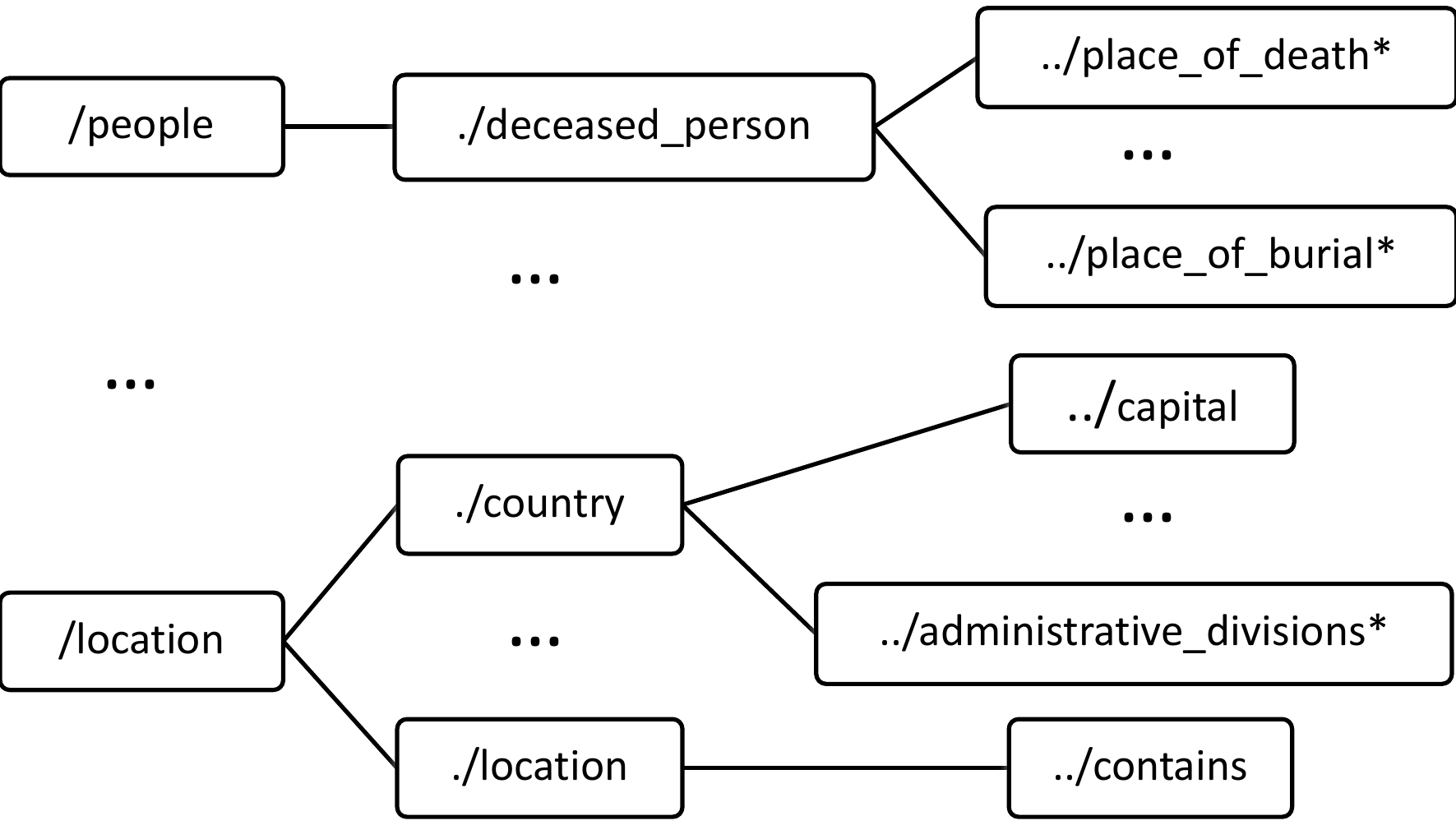}
    }
	\subfigure
	{
		\includegraphics[width=0.253\textwidth]{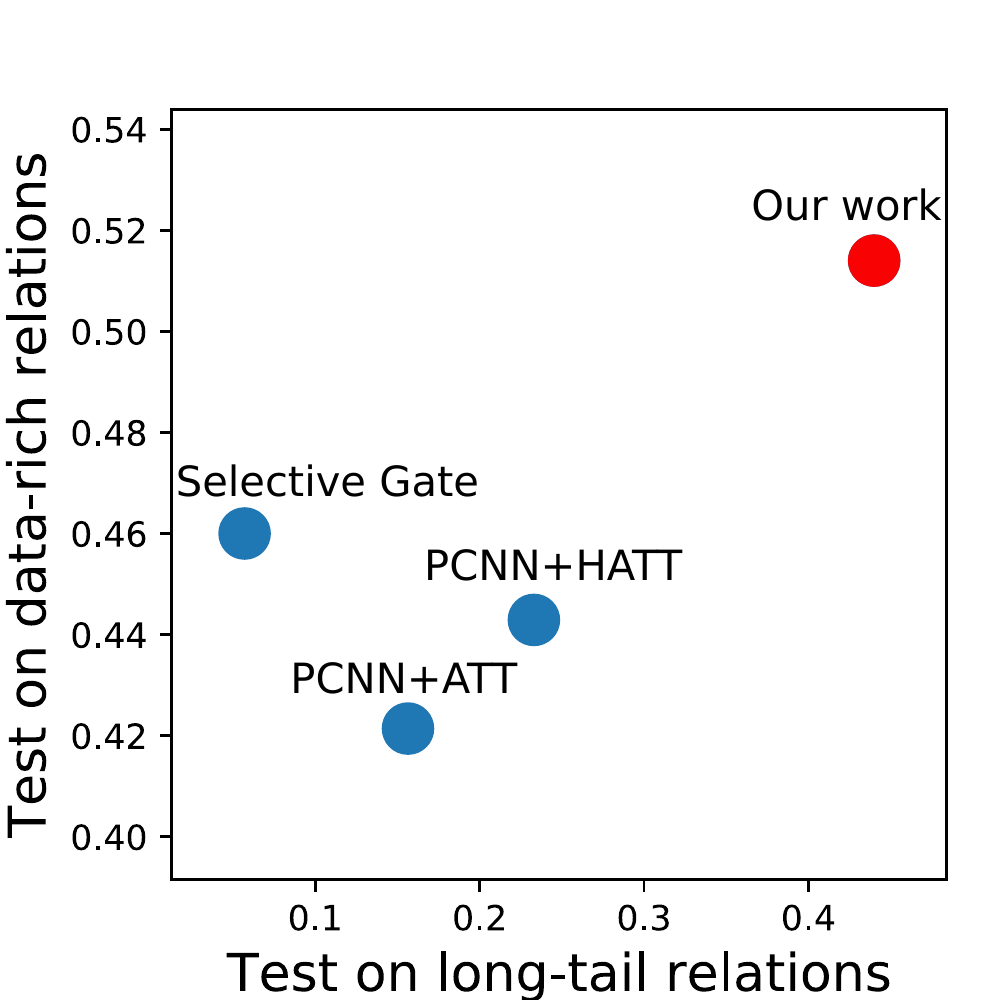}
	}
	\caption{ \small
	\textbf{Left}: Label frequency distribution of relations without \textit{NA} in NYT dataset. Here the criterion being a long-tail relation is the number of corresponding training sentences is less than 1000.
	\textbf{Middle}: Relation hierarchies in Freebase knowledge graph, where relation with ``$*$'' is long-tail. Through the common high-level relation, it can be exploited that 1) multiple semantic-related low-level relations complement each other and 2) semantic knowledge is transferred from data-rich to long-tail relations.
	\textbf{Right}: Empirical AUC results of competitive approaches on data-rich and long-tail test subsets.
	}
	\label{fig:longtail}
\end{figure}

Second, the \textit{long-tail problem} is caused by using a knowledge graph as distant supervision to auto-label a domain-specific corpus, where the knowledge graph usually suffers from long-tail relations. For example, to build NYT dataset, applying Freebase to a news corpus, New York Times, leads to $\sim$70\% of relations long-tail, as illustrated in Figure \ref{fig:longtail} (left). This problem seriously disrupts the data balance and thus becomes one of the main barriers of improvement. 

%  via the relation hierarchies of a knowledge graph containing semantic overlap
To alleviate the long-tail problem, two approaches \cite{han2018hierarchical,zhang2019long} naturally share the knowledge from data-rich relations to the long-tail ones when those relations have semantic overlap. This semantic overlap or relatedness is usually stored in the relation hierarchies of a knowledge graph, e.g., Freebase in Figure \ref{fig:longtail} (middle). 
Specifically, these approaches extend selective attention \cite{lin2016neural} by introducing the embeddings of high-level (i.e., coarse-grained) relations as complements to original low-level (i.e., fine-grained) relations. As such, the high-level relation embeddings are used as queries of selective attention to derive extra bag-level features. 
To learn the relation embeddings, Han et al.~\shortcite{han2018hierarchical} randomly initialize them followed by supervised learning in end-to-end fashion, whereas Zhang et al.~\shortcite{zhang2019long} combine the embeddings from both TransE \cite{bordes2013translating} pre-trained on Freebase and graph convolutional network \cite{defferrard2016convolutional} applied to the relation's hierarchies. 

Despite proven to improve overall and long-tail performance, they also post two issues: 
1) Limited by selective attention framework, the relation embeddings are only used as the attention's queries and thus not well-exploited to share knowledge. 
2) Despite the capability in mitigating the long-tail problem, graph embeddings pre-trained on large-scale knowledge graph are time-consuming and not always off-the-shelter, hence at the cost of practicability. 

In this work, we propose a novel neural network, named as \textbf{Co}llaborating \textbf{R}elation-augmented \textbf{A}ttention (CoRA), to tackle distantly supervised relation extraction, where no external knowledge is introduced and the relation hierarchies are fully utilized to alleviate the long-tail problem. 
Specifically, as an alternative to selective attention framework, we first propose a base model, \textit{relation-augmented attention}, operating at bag level to minimize the effect of wrong labeling, where the relation-augmenting process is fulfilled by a sentence-to-relation attention. 
Empowered by the base model, we then leverage the high-level relations for collaborating features in light of the relation hierarchies. Besides a further relief of wrong labeling, such features facilitate knowledge transfer among the low-level relations inheriting a common high-level relation. 

Intuitively, selective attention and its hierarchical extensions learn relation label embeddings to score each sentence in a bag. 
In contrast, the proposed relation-augmented attention network achieves the same goal via a memory network-like structure: sentences equipped with relation features are passed into an attention-pooling (i.e., a kind of self-attention \cite{lin2017structured}) for bag-level representations. 
Our method is especially effective when extended to multi-granular relations -- the features are enriched by cross-relation sharing, which hence benefits long-tail relations. As shown in Figure \ref{fig:longtail} (right), our proposed approach achieves consistent outstanding performance on both data-rich and long-tail relations. 

We use two objectives to jointly train the CoRA. The first is predicting the relation label at bag level, which is the goal of relation extraction. 
As auxiliary objective, the second is guiding the model to equip each sentence with correct multi-granular relation embeddings during the augmenting process. It aims to boost downstream attention-pooling and is fulfilled by applying the multi-granular labels to the sentence-to-relation attention during training.

Our main contributions are summarized as: 
\begin{itemize}
\setlength{\parsep}{0pt}
\setlength{\parskip}{0pt}
    \item  We propose a base model, named relation-augmented attention network, to handle wrong labeling problem in multi-instance learning. 
    \item We then propose to extend the base model with the relation hierarchies, called CoRA, to further promote the performance on long-tail relations.
    \item We evaluate CoRA on the popular benchmark dataset NYT and set state-of-the-art results in Precision@N, AUC and Hits@K. We also verify its capability in alleviating both wrong labeling and long-tail problems via insightful analyses. The source code of this work is released at \url{https://github.com/YangLi1221/CoRa}.
\end{itemize}

\begin{figure*}[t]
    \centering
    \includegraphics[width=0.94\textwidth]{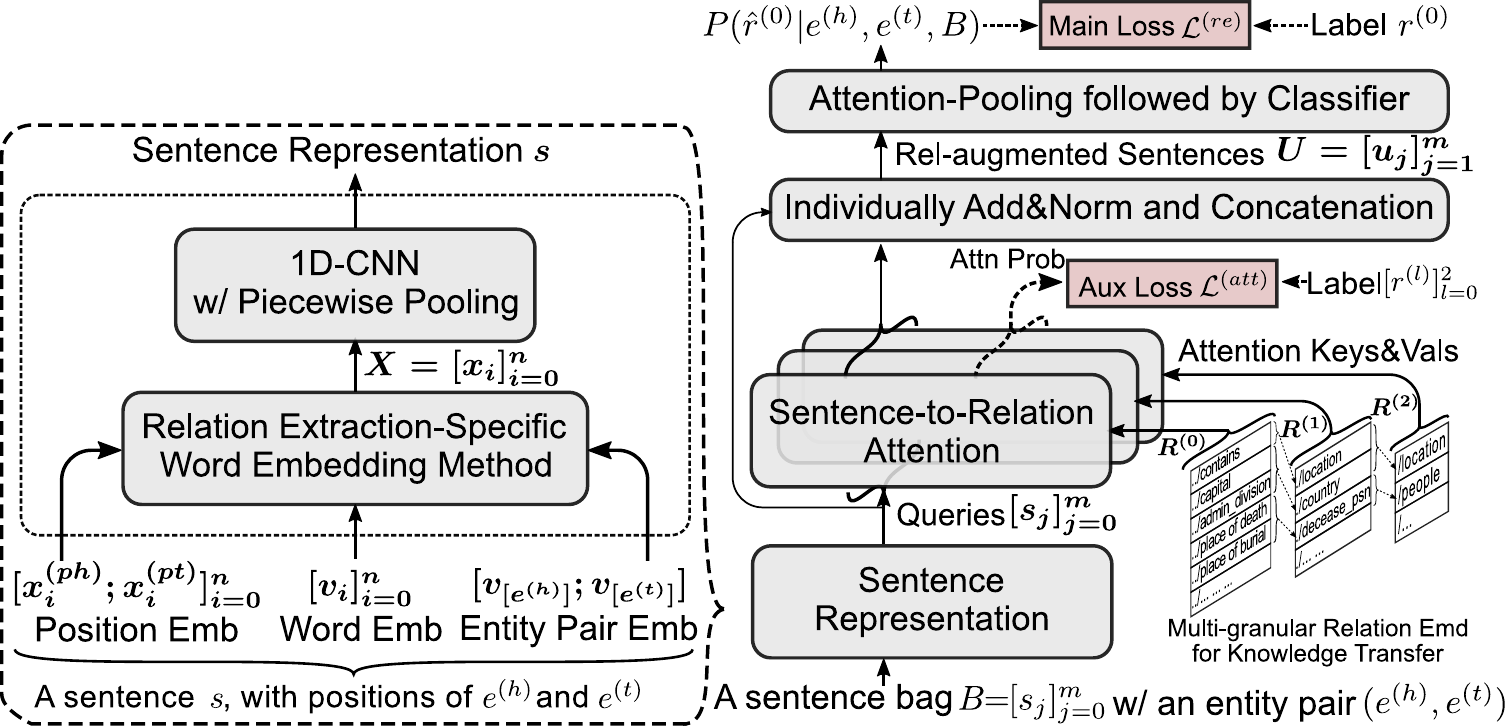}
    \caption{ Our proposed \textbf{Co}llaborating \textbf{R}elation-augmented \textbf{A}ttention (CoRA) Network, where the \textit{right part} is the main structure while the \textit{left part} is a sentence embedding method for relation extraction. The illustrated relations and their hierarchies are based on NYT dataset where $M=2$ in Eq.(\ref{eq:rel_aug_repre_cora}).
    }
    \label{fig:model}
\end{figure*}

\section{Proposed Approach}
This section begins with a definition of distantly supervised relation extraction with multi-granular relation labels. Then an embedding method is introduced to represent sentences. 
Lastly, our base model and its hierarchical extension are presented to handle wrong labeling and long-tail relations. 
% Next, relation-augmented attention is proposed to better tackle the multi-instance learning problem, and the proposed approach is further extended to collaborating relation-augmented attention for alleviating long tail problems when integrated with the hierarchies. 
An illustration of the model is shown in Figure \ref{fig:model}. 

\subsection{Task Definition}
Given a bag of sentences $B = \{s_1, \dots, s_m\}$ in which each sentence contains a pair of head $e^{(h)}$ and tail $e^{(t)}$ entities in common, the distant supervision \cite{mintz2009distant} assigns this bag with a relation label $r^{(0)}$ according to the entity pair in a knowledge graph. The goal of relation extraction is to predict the relation label $\hat r^{(0)}$ of an entity pair based on the corresponding sentence bag when the pair is not included in the knowledge graph. 
As following the hierarchical setting \cite{han2018hierarchical,zhang2019long}, labels of coarse-grained relations, [$r^{(1)}$, \dots, $r^{(M)}$], can be used to share knowledge across relations. 

\subsection{Sentence-Level Representation} \label{sec:sent_repre}
To embed each sentence $s_j$ in a bag $B = \{s_1, \dots, s_m\}$ into latent semantic space, we derive a sentence representation from three kinds of features, including word embedding \cite{mikolov2013distributed}, position embedding \cite{zeng2015distant} and entity embedding \cite{li2020self}. The integration of them has been proven crucial and effective to relation extraction by previous work \cite{li2020self}. In the following, we omit the index of a sentence, $j$, for a clear elaboration. Basically, a sentence $s$ is first tokenized into a sequence of $n$ words, $s = [w_1, \dots, w_n]$, then a word2vec method \cite{mikolov2013distributed} is used to transform the discrete tokens into low-dimensional, real-valued vector embeddings, i.e., $\mV = [\vv_1, \dots, \vv_n]\in\R^{d_w\times n}$. 
%where $n$ is the sequence length. 

\paragraph{Word Embedding. }
On the one hand, \textit{position-aware embedding} offers rich positional information for downstream modules \cite{zeng2014relation}. For $i$-th word, the relative position is represented as the distances from the word to head $e^{(h)}$ and tail $e^{(t)}$ entities respectively. Two integer distances are then transformed into low-dimensional vectors, $\vx_i^{(ph)}$ and $\vx_i^{(pt)}\in\R^{d_p}$, by a learnable weight matrix. Consequently, a sequence of position-aware embeddings is denoted as $\mX^{(p)} = [\vx_1^{(p)}, \dots, \vx_n^{(p)}]\in\R^{(d_w+2d_p)\times n}$ where $\vx_i^{(p)} = [\vv_i, \vx_i^{(ph)}; \vx_i^{(pt)}]$. $[;]$ denotes the operation of vector concatenation. 
On the other hand, \textit{entity-aware embedding} is also crucial since the goal of relation extraction is to discriminate the relation between two entities. The embedding of head or tail entity is represented by the corresponding word embedding. Note that each entity is one entry in the vocabulary of word embedding even if it is usually composed of multiple words. Hence, a sequence of entity-aware embeddings is denoted as $\mX^{(e)} = [\vx_1^{(e)}, \dots, \vx_n^{(e)}]\in\R^{3d_w\times n}$ where $\vx^{(e)} = [\vv_i, \vv_{[e^{(h)}]}; \vv_{[e^{(t)}]}] \in\R^{3d_w}$. 
To integrate the embeddings above, a position-wise gating procedure is employed by following Li et al.~\shortcite{li2020self}. That is,
\begin{align}
	&\mA^{(e)} = \mysigmoid(\lambda \cdot (\mW^{(g1)} \mX^{(e)} + \vb^{(g1)})), \label{eq:hyperparameter} \\ 
	&\tilde{\mX}^{(p)} = \tanh(\mW^{(g2)}\mX^{(p)} + \vb^{(g2)}), \\
	&\mX = \mA^{(e)} \circ \mX^{(e)} + (1-\mA^{(e)}) \circ \tilde{\mX}^{(p)},
\end{align}
where ``$\circ$'' denotes element-wise product $\mW^{(g1)}\in\sR^{d_x\times{3d_w}}$ and $\mW^{(g2)}\in\sR^{d_x\times{(d_w + 2d_p)}}$ are learnable parameters, $\lambda$ is a hyper-parameter to control smoothness, and $\mX = [\vx_1, \ldots, \vx_n]\in\sR^{d_x\times n}$ is the resulting sequence of word embeddings specially for relation extraction.

\paragraph{Piecewise Convolutional Neural Network. }
As a common practice in distantly supervised relation extraction, piecewise convolutional neural network (PCNN) \cite{zeng2015distant} is used to generate contextualized representations over an input sequence of word embeddings. 
Compared to the typical 1D-CNN with max-pooling \cite{zeng2014relation}, piecewise max-pooling has the capability to capture the structure information between two entities by considering their positions. 
Specifically, 1D-CNN \cite{kim2014convolutional} is first invoked over the input sequence for contextualized representations. Then a piecewise max-pooling performs over the output sequence to obtain sentence-level embedding. % : the sequence is divided into three segments w.r.t. indices of head and tail entities, and mean-pooling is apply to each segments. 
These steps are written as
\begin{align}
    &\mH = [\vh_1, \dots, \vh_n] = \onedcnn(\mX; \mW^{(c)}, \vb^{(c)}) \in \R^{d_c \times n}, \\
    &\vs = \tanh([\pool(\mH^{(1)});\pool(\mH^{(2)});\pool(\mH^{(3)})]),
\end{align}
where $\mW^{(c)}\in\R^{d_c \times Q \times d_x}$ is a conv kernel with window size of $Q$. $\mH^{(1)}$, $\mH^{(2)}$ and $\mH^{(3)}$ are three consecutive parts of $\mH$, obtained by dividing $\bm{H}$ w.r.t. indices of head $e^{(h)}$ and tail  $e^{(t)}$ entities. % according to the positions of head and tail entities. 
Consequently, $\vs\in\sR^{d_h}$, where $d_h = 3d_c$, is the resulting sentence-level representation.

\subsection{Relation-Augmented Attention Network} \label{sec:method_ra}
Due to the effectiveness of selective attention \cite{lin2016neural} in multi-instance learning, most recent works employ the selective attention as the baseline and then propose own approaches for improvements in wrong labeling and/or long-tail relations. 
However, selective attention gradually becomes a bottleneck of performance improvement. 
For example, Li et al.~\shortcite{li2020self} find using simple gating mechanism to replace selective attention further alleviates wrong labeling problem and significantly promotes extracting results. 
Intuitively, on the one hand, employing the basic PCNN and vanilla attention mechanism inevitably limits the expressive power of this framework and thus sets a barrier. On the other hand, the relation embeddings, similar to label embeddings \cite{bengio2010label}, are crucial to distant supervision relation extraction, but are only used as attention query to score a sentence and thus not well-exploited.

In contrast, we aim to augment each sentence in a bag with the relation embeddings by a sentence-to-relation attention, and pass the relation-augmented representations of a bag's sentences into an attention-pooling module. 
The attention-pooling, a kind of self-attention \cite{lin2017structured,shen2018biblosan,shen2018disan}, is used to derive an accurate bag-level representation for relation classification. 
In details, we first define a relation embedding matrix $\mR^{(0)}\in\R^{d_h\times N^{(0)}}$ where $d_h$ denotes the size of hidden states and $N^{(0)}$ denotes the number of distinct relations $r^{(0)}$ in a distantly supervised relation extraction task. 
Then, we formulate a sentence-to-relation (sent2rel) attention as opposed to selective attention, which aims at augmenting sentence representation from \S\ref{sec:sent_repre} with relation information. The sentence representation $\vs$ is used as a query to attend the relation embedding matrix $\mR^{(0)}$ via a dot-product compatibility function: % It is written as
\begin{align}
    &\bm{\alpha}^{(0)} = \softmax(\vs^T\mR^{(0)}), \label{eq:sent2rel_attn} \\ 
    &\vc^{(0)} = \mR^{(0)} \bm{\alpha},
\end{align}
where $\softmax(\cdot)$ denotes a normalization function along last dimension and $\vc^{(0)}$ is the resulting relation-aware representation corresponding to the sentence $\vs$. Then we merge the relation-aware representation $\vc^{(0)}$ into original sentence representation $\vs$ by an element-wise gate mechanism with residual connection \cite{he2016deep} and layer normalization \cite{ba2016layer}, i.e.,
\begin{align}
    &\bm{\beta}^{(0)} = \mysigmoid(\mW^{(g)} [\vs; \vc^{(0)}] + \vb^{(g)}), \\
    &\tilde \vu^{(0)} = \bm{\beta}^{(0)} \circ \vs + (1-\bm{\beta}^{(0)}) \circ \vc^{(0)}, \\
    &\vu^{(0)} = \layernorm(\vs + \mlp(\tilde \vu^{(0)})), \label{eq:sub_rel_aug_repre}
\end{align}
where $\mlp(\cdot)$ denotes a multi-layer perceptron to increase nonlinearity. 
Finally, we define relation-augmented sentence representation in our base model as
\begin{align}
    \vu \coloneqq \vu^{(0)}. \label{eq:rel_aug_repre}
\end{align}

Next moving to multi-instance learning, we put each sentence back to its bag $B = \{s_1, \dots, s_m\}$ so the bag of sentences with relation-augmentation is represented as $\mU = [\vu_1, \dots, \vu_m]\in\R^{d_h\times m}$. Differing from selective attention framework, our sentence representations are augmented by the relation embeddings as elaborated above. Hence, we straightforwardly introduce an attention-pooling module to derive a bag-level representation denoising from the wrongly-labeled sentences. Specifically, the attention-pooling learns to assign each sentence with an importance score according its representation. Then it performs a weighted sum over a bag of sentence representations, where the weights are proportional to their scores. This attention is formulated as
\begin{align}
    \vb = \mU \softmax(\vw^T \mU),
\end{align}
where $\vw$ is a learnable weight vector, and $\vb$ denotes the resulting bag-level representation. 
Lastly, an MLP is used to obtain a categorical distribution over all relations as bag-level prediction:
\begin{align}
    \vp = P(\hat r^{(0)}|e^{(h)}, e^{(t)}, B) \coloneqq \mlp(\vb) \in\R^{N^{(0)}}. \label{eq:bag_level_predict}
\end{align}

\subsection{Collaborating Relation-Augmented Attention Network} \label{sec:method_cora}

Beyond only fine-grained relations used above, high-level relation embeddings as hierarchical knowledge can collaborate with the low-level embeddings to boost the performance by alleviating long-tail problem \cite{han2018hierarchical,zhang2019long}.
Intuitively, a high-level relation, shared crossing several low-level relations, is used to represent common knowledge of low-level relations. 
Therefore, via the common high-level relation, 1) several low-level long-tail relations with semantic overlap mutually benefit each other, and 2) the semantic knowledge is easily transferred from data-rich relations to long-tail ones. 
These common knowledge is implicitly utilized to distinguish the coarse-grained relation of a bag and thus benefits the final relation prediction. 
With the relation-augmented sentence representation further enriched via collaborating, we name it as \textbf{Co}llaborating \textbf{R}elations-augmented \textbf{A}ttention (CoRA). 

Empowered by non-trivial structure design of our base model, high-level relation embeddings can be easily integrated into the base model by re-defining Eq.(\ref{eq:rel_aug_repre}). In particular, given the coarse-grained relation labels from low to high level, i.e., $[r^{(1)}, \dots, r^{(M)}]$, we define a list of relation embedding matrices $[\mR^{(1)}, \dots, \mR^{(M)}]$ in addition to $\mR^{(0)}$ defined in last section. 
With these relation embedding matrices, we individually generate their corresponding relation-augmented sentence representations, i.e., $[\vu^{(1)}, \dots, \vu^{(M)}]$, via the same procedure defined in Eq.(\ref{eq:sent2rel_attn} -- \ref{eq:sub_rel_aug_repre}) of \S\ref{sec:method_ra}. Then, we concatenate $[\vu^{(1)}, \dots, \vu^{(M)}]$ in conjunction with $u^{(0)}$ to re-formulate Eq.(\ref{eq:rel_aug_repre}) as
\begin{align}
    \vu \coloneqq [\vu^{(0)}; \vu^{(1)}; \dots, \vu^{(M)}] \in\R^{(1+M)d_h}. \label{eq:rel_aug_repre_cora}
\end{align}
The following procedure is identical to that in base model elaborated above, except that the learnable weight matrices are up-scaled linearly with the depth of relation hierarchies. 

\subsection{Training Objectives} \label{sec:train_objs} 
The \textit{main objective} for relation extraction is defined to minimize a cross-entropy loss, i.e., 
\begin{align} % \dfrac{1}{|\gD|}
    \gL^{(re)} = - \dfrac{1}{|\gD|} \sum\nolimits_{B\in\gD} \log P(\hat r^{(0)} = r^{(0)}|e^{(h)}, e^{(t)}, B),
\end{align}
where $\gD$ is the training set consisting of sentence bags. Besides, an \textit{auxiliary objective} guides sentence-to-relation attention modules to augment each sentence with correct relation embeddings. This is critical to perform downstream attention-pooling and overcome the challenges presented by distant supervision. Given the sent2rel attention score $\bm{\alpha}^{(l)}$ and relation label $r^{(l)}$ at an arbitrary $l$ level, the loss function to achieve this objective is defined as
\begin{align}
    % \gL^{(att)} = - \dfrac{1}{|\gD| \cdot |B|\cdot (1+M)} \sum_{B\in\gD} \sum_{s\in B} \sum_{l=0}^{M} \log \bm{\alpha}^{(l)}_{[r^{(l)}]}. 
    \gL^{(att)} = - \dfrac{1}{|\gD| \cdot |B|\cdot (1+M)} \sum\nolimits_{B\in\gD} \sum\nolimits_{s\in B} \sum\nolimits_{l=0}^{M} \log \bm{\alpha}^{(l)}_{[r^{(l)}]}.
    \label{eq:loss_fn_att}
\end{align}
where $M=0$ for the base model in \S\ref{sec:method_ra}, where $M>0$ for CoRA in \S\ref{sec:method_cora}. Finally, we optimize the proposed model by jointly minimizing the two loss functions above, i.e., $\gL = \gL^{(re)} + \gL^{(att)}.$
% \begin{align}
%     \gL = \gL^{(re)} + \gL^{(att)}. \label{eq:final_loss_fn}
% \end{align}

\section{Experiments}
% We evaluate our proposed network on a popular benchmark dataset and compare it with recently-proposed competitors. 
% And we also conduct ablation studies and comprehensive analyses for insights into our proposed network.
We evaluate our proposed network on a popular benchmark dataset and conduct several analyses for insights into our proposed model.
%We compare our proposed network with recently-proposed competitors on a popular benchmark dataset with ablation studies and comprehensive analyses for insights into our proposed network.

\paragraph{Dataset and Evaluation Metrics.} By following previous works \cite{zeng2015distant,lin2016neural,han2018hierarchical}, we employ the only popular distantly supervised relation extraction dataset, New York Times (NYT) dataset \cite{riedel2010modeling}. It contains 53 distinct relations which includes a \textit{NA} class denoting the relation between the entity pair is unavailable. And it consists of 570K and 172K sentences in training and test sets respectively. Two metrics, 1) area under precision-recall curve (AUC) and 2) top-n precision (P@N) are usually used to measure the effectiveness. We also use Hits@K for long-tail relations by following Zhang et al.~\shortcite{zhang2019long}.

\paragraph{Setups.}
Following previous works, $d_w$, $d_p$, $d_x$, $d_c$, $d_h$ and $Q$ are 50, 5, 150, 230, 690 and 3 respectively. $\lambda$ in Eq.(\ref{eq:hyperparameter}) is 0.05. NYT offers two more high-level (coarse-grained) relations (i.e., $M=2$), and the numbers of distinct relations at three levels are 53, 36 and 9. 
During training, we use minibatch SGD \cite{zeiler2012adadelta} with Adam \cite{kingma2014adam} optimizer. The learning is $0.1$, batch size is 160, dropout probability is set to 0.5, weight decay of L2 regularization is $10^{-5}$.

\paragraph{Comparative Approach.} We compare the proposed approach with extensive previous works that are summarized as follows. A model with ``$*$'' denotes it is proposed for the long-tail problem. 
\begin{itemize}
\setlength{\parsep}{0pt}
\setlength{\parskip}{0pt}
	\item \textbf{PCNN+ATT} \cite{lin2016neural} proposes a selective attention to alleviate wrong labeling. 
	\item \textbf{PCNN+HATT}$^*$ \cite{han2018hierarchical} employs hierarchical attention to exploit the relations.
	%, which shares a similar motivation with ours.
	\item \textbf{PCNN+BAG-ATT} \cite{ye2019distant} proposes intra-bag and inter-bag attentions to handle wrongly-labeled sentences at sentence level and bag level respectively.
    \item \textbf{PCNN+KATT}$^*$ \cite{zhang2019long} integrates externally pre-trained graph embeddings with relation hierarchies for long-tail relations. Note, standard AUC and P@N values are not available in the paper while only Hits@K is defined and reported for long-tail settings. 
    %\item \textbf{RELE} \cite{hu2019improving} proposes a multi-layer attention-based model with joint label embedding to leverage the information from knowledge graphs and entity descriptions. 
    \item \textbf{SeG} \cite{li2020self} focuses on one-sentence bags and proposes selective gate mechanism.
    %a light-weight neural network with selective gate mechanism to handle them. 
\end{itemize}

\begin{table*}[t]\small
	\centering
    \setlength{\tabcolsep}{1.5pt}
	\begin{tabular}{lcccccccccccc|c}
		\hline
		\multirow{2}{*}{ \textbf{P@N (\%)} }&\multicolumn{4}{c}{\bf One}&\multicolumn{4}{c}{\bf Two}&\multicolumn{4}{c|}{\bf All}&\multirow{2}{*}{\textbf{AUC}}\\
		\cline{2-13} 
		 &100&200&300&Mean&100&200&300&Mean&100&200&300&Mean&\\
		\hline
		\multicolumn{13}{l}{\textit{Comparative Approaches} } & \\
		\hline
		CNN+ATT \cite{lin2016neural} &72.0&67.0&59.5&66.2&75.5&69.0&63.3&69.3&74.3&71.5&64.5&70.1&0.35\\
		PCNN+ATT \cite{lin2016neural} &73.3&69.2&60.8&67.8&77.2&71.6&66.1&71.6&76.2&73.1&67.4&72.2&0.39\\
		PCNN+HATT \cite{han2018hierarchical} &84.0&76.0&69.7&76.6&85.0&76.0&72.7&77.9&88.0&79.5&75.3&80.9&0.42\\
		PCNN+BAG-ATT \cite{ye2019distant}  &86.8&77.6&73.9&79.4&91.2&79.2&75.4&81.9&91.8&84.0&78.7&84.8&0.42\\
		%RELE \cite{hu2019improving} &87.0&76.6&67.0&76.8&88.0&73.3&63.2&78.7&88,0&78.6&69.8&78.8&-\\ 
		SeG \cite{li2020self} &\textbf{94.0}&89.0&\textbf{85.0}&\textbf{89.3}&91.0&89.0&\textbf{87.0}&89.0&93.0&90.0&86.0&89.3&0.51\\
		\hline
		\textbf{CoRA} (\textit{ours}) &\textbf{94.0}&\textbf{90.5}&82.0&88.8&\textbf{98.0}&\textbf{91.0}&86.3&\textbf{91.8}&\textbf{98.0}&\textbf{92.5}&\textbf{88.3}&\textbf{92.9}&\textbf{0.53}\\
		\midrule [0.2ex]
		\multicolumn{13}{l}{\textit{Ablations} } & \\
		\hline
		Base$^*$ (CoRA w/o Collaborating) &90.0&89.0&85.3&88.1&93.0&90.0&85.3&89.4&93.0&90.5&87.0&90.2&0.52\\
		Base w/o Ent Emb in \S\ref{sec:sent_repre} &83.0&74.0&69.3&74.5&84.0&81.0&72.3&79.1&85.0&80.0&73.3&79.4&0.45\\
		Base w/o Sent2rel Attention in \S\ref{sec:method_ra} &83.0&74.0&66.6&74.5&82.0&79.0&68.3&76.5&84.0&79.5&73.0&78.8&0.43\\
		Base w/o Attention-pooling in \S\ref{sec:method_ra} &90.0&87.0&84.0&87.0&93.0&88.0&85.0&88.7&94.0&88.5&86.0&89.5&0.52\\
		Base w/o Aux Obj $\gL^{(att)}$ in Eq.(\ref{eq:loss_fn_att}) &80.0&70.0&65.7&71.9&83.0&74.0&68.0&75.0&85.0&80.0&70.3&78.4&0.41\\
		\hline
	\end{tabular}
	\caption{Model Evaluation and ablation study on NYT. ``P@N'' (top-n precision) denotes precision values for the entity pairs with top-100, -200 and -300 prediction confidences by randomly keeping one/two/all sentence(s) in each bag. $^*$Base model denotes relation-augmented attention network where $M=0$.}
	\label{tab:eval_on_benchmark}
\end{table*}

\subsection{Evaluation on Benchmark}
\begin{figure}[t]
	\centering
	\subfigure
	{
		\includegraphics[width=0.315\textwidth]{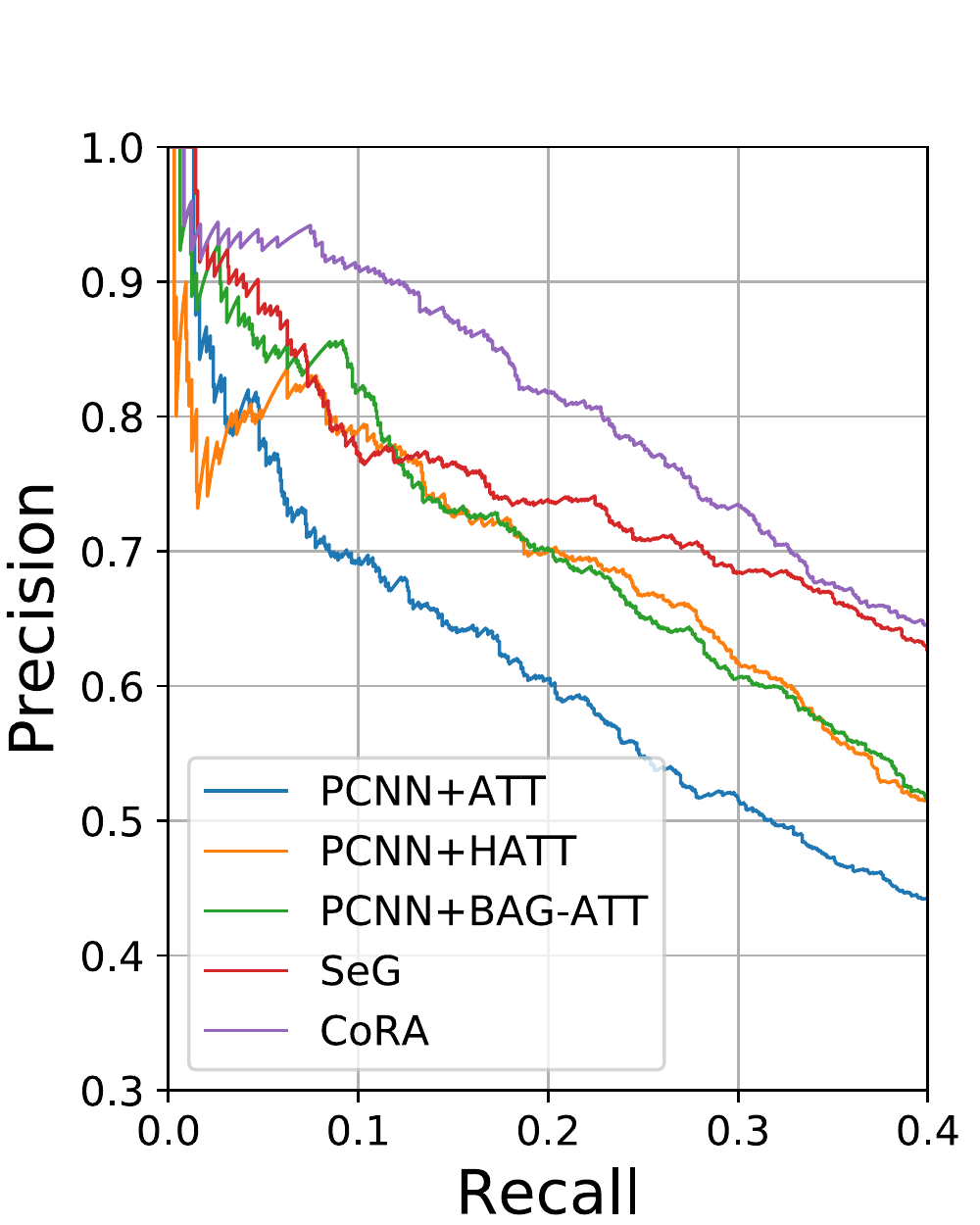}
    }
    \subfigure
    {
        \includegraphics[width=0.315\textwidth]{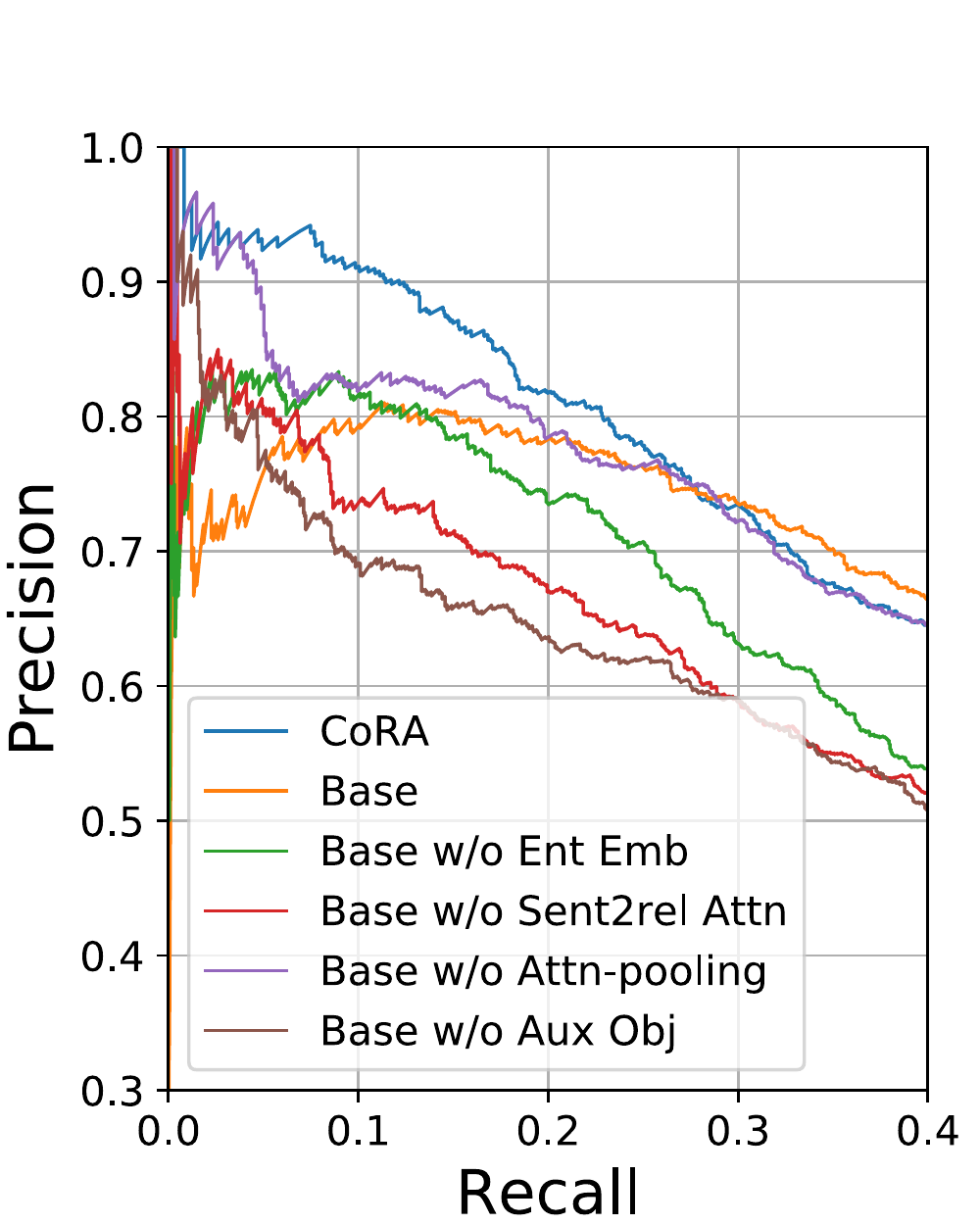}
    }
    \subfigure
    {
        \includegraphics[width=0.315\textwidth]{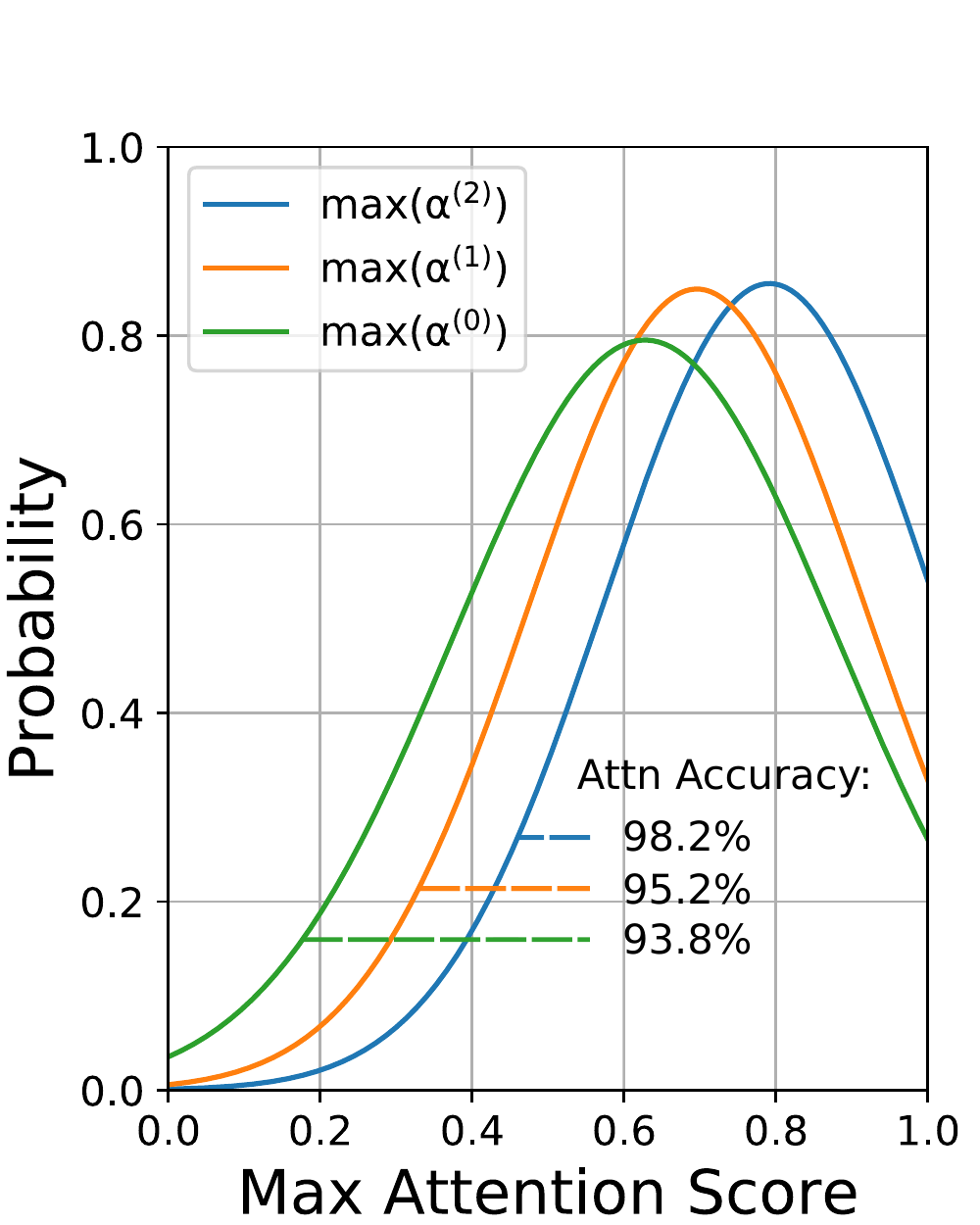}
    }
	\caption{\textbf{Left}: Precision-recall (PR) curves on NYT for model comparison. \textbf{Middle}: PR curves for ablation study. \textbf{Right}: Probability (normal) distribution of maximum attention score, $\max(\bm{\alpha}^{(l)}))$, in sent2rel attention, where attention accuracy is whether the the max score $\max(\bm{\alpha}^{(l)})$ corresponds to $r^{(l)}$.
	}
	\label{fig:pr_curve}
\end{figure}

As shown in Table \ref{tab:eval_on_benchmark} and Figure \ref{fig:pr_curve} (left), we compare our CoRA with previous competitive approaches on the distantly supervised relation extraction benchmark in terms of top-n precision, AUC and PR curve. 
Specifically, CoRA significantly outperforms selective attention baseline, i.e., PCNN+ATT. It also surpasses selective gate framework that shows inferior performance on long-tail relations as in Figure \ref{fig:longtail} of \S\ref{sec:intro}.
In addition, compared to PCNN+HATT also utilizing relation hierarchies, CoRA is able to achieve much better results in both P@N and AUC. 

\subsection{Ablation Study}
To further evaluate the effectiveness of each module in the proposed framework, we conduct an extensive ablation study in the bottom of Table \ref{tab:eval_on_benchmark} and Figure \ref{fig:pr_curve} (middle). Since the performance drop is consistent in P@N and AUC, we mainly use AUC as the metric to perform following study. 
Compared to CoRA, base model without relation collaborating features only shows marginal precision drop when recall $>$ 0.3 in PR-curve, but a significant drop on long-tail relations (detailed in the next section). Also, as an alternative to selective attention, our base model outperforms PCNN+ATT by a large margin. 
Then, removing simple entity embeddings in \S\ref{sec:sent_repre} leads to remarkable degeneration, verifying its importance. It is also rational to compare PCNN+ATT with ``Base w/o Ent Emb'' (+0.06 AUC) to demonstrate our relation-augmented framework is indeed better than selective attention. 
Then, removing ``Sent2rel Attention'', ``Attention-pooling'' and ``Aux Obj'' reduce the AUC by 0.10, 0.01 and 0.12 respectively. %Note, ``w/o attention pooling'' means applying mean-pooling to a bag of sentence representations for bag-level representation. 

\begin{table}[t]
    \centering
    \begin{tabular}{lcccccc}
        \toprule
        \multicolumn{1}{l}{\bf \# Training Instance}&\multicolumn{3}{c}{\bf \textless100}&\multicolumn{3}{c}{\bf \textless200}\\ \midrule
        \multicolumn{1}{l}{\textbf{Hits@K}  (Macro)}&\textbf{10}&\textbf{15}&\textbf{20}&\textbf{10}&\textbf{15}&\textbf{20}\\
        \midrule
         PCNN+ATT \cite{lin2016neural} & \textless5.0&7.4&40.7&17.2&24.2&51.5 \\
         PCNN+HATT \cite{han2018hierarchical} & 29.6&51.9&61.1&41.4&60.6&68.2 \\
         PCNN+KATT \cite{zhang2019long} & 35.3&62.4&65.1&43.2&61.3&69.2 \\ 
        \midrule
         CoRA & \textbf{66.6}&\textbf{72.0}&\textbf{87.0}&\textbf{72.7}&\textbf{77.3}&\textbf{89.4} \\
        \midrule
         Base & 33.3&44.4&66.6&45.5&54.5&72.7\\
        %  Basic model w/o Ent & 8.3&11.11&45.83&21.4&23.8&53.6\\
         Base w/o Aux Obj & 18.5&44.4&61.1&33.3&54.5&68.1\\
         %Basic model w/ mean pooling & 48.6&60.8&70.3&56.0&65.0&75.8\\
         Base w/o Sent2rel Attention & 5.0&33.3&61.1&22.7&45.5&68.1\\
        \bottomrule
    \end{tabular}
    \caption{Hits@K (Macro) on the relations whose number of training instance $<$ 100/200.
    ``Hits@K'' denotes whether a test sentence bag whose gold relation label $r^{(0)}$ falls into top-$K$ relations ranked by their prediction confidences. 
    ``Macro'' denotes macro average is applied regarding relation labels.
    %``Macro'' is an average is applied to all the Hits@K results that have been averaged in all relations.  
    % $^*$The reported numbers are obtained by our reimplementation.
    }
    \label{tab:hits@k}
\end{table}

\subsection{Evaluation on Long-Tail Relations}
To prove the capability of CoRA in handling long-tail relations, we conduct an evaluation solely on long-tail relations. Our evaluation setting is identical to \cite{han2018hierarchical,zhang2019long}, where Hits@K (Macro) is used to represent statistical performance on long-tail relations. 
As shown in Table \ref{tab:hits@k}, we compare CoRA with competitors and our base models. 
It is observed that, CoRA improves the performance on long-tail relations by a large margin and delivers a new state-of-the-art results. Compared to previous works (PCNN+HATT/+KATT) that also leverage the relation hierarchies, our relation-augmented attention (Base) without any hierarchy even gets competitive results, not to mention pre-trained graph embeddings used in PCNN+KATT. Further comparing our base model with selective attention (PCNN+ATT), the huge performance gap demonstrates the advantages of our framework in handling both wrong labeling and long-tail relations. 
Finally, as shown in the table's last row, removing the proposed sent2rel attention leads to significant decrease, which emphasizes its importance for long-tail relations.

\subsection{Analysis and Case Study}

% In this section, we provide insights into our proposed approach from multiple perspectives. 

\paragraph{Distributions of Sent2rel Attention Scores. }
Sent2rel attention used to incorporate multi-granular relation embeddings is an essential module in CoRA, so its normalized attention scores (i.e., attention probabilities) derived from Eq.(\ref{eq:sent2rel_attn}) are critical to measure the knowledge transfer crossing relations. We show a probability distribution of maximum attention score in Figure \ref{fig:pr_curve} (right). 
Obviously, a high-level sent2rel attention tends to produce larger maximum attention score and more accurate attention target. It is easily inferred that, 1) accurate attention at high-level promotes the knowledge transfer through the relation hierarchies, and 2) attention probability distribution is more smooth at low-level to further boost embedding sharing crossing relations. 
To dig this out, in Table \ref{tab:casestudy}, we conduct a case study by showing top attention scores at all three relation levels. 
% Both the sentences express a long tail relation, ``\textit{/business/company/founders}'', in NYT dataset. 
It is observed that attention scores and the corresponding relations are intuitively consistent with the analyses above. One exception is that \textit{NA} class appears to be assigned with high attention score at low-level sent2rel attention, which indirectly explains 1) our base model w/o collaborating relation features only delivers inferior performance and 2) sent2rel attention for low-level relations are inaccurate. 

\begin{table*}[t]
    \small
    \centering
    \begin{tabular}{p{80pt}|p{20pt}|p{90pt}|p{20pt}|p{120pt}|p{20pt}}
    \toprule
    % \multicolumn{6}{c}{\centering Bag-level Relation: \textit{/business/company/founders}} \\
    % \hline
    \multicolumn{6}{l}{\textbf{Example Sentence 1:} \textit{Muhammad\_yunus}, who won the nobel peace prize, last year, demonstrated with \textit{grameen\_bank},} \\
    \multicolumn{6}{l}{\hspace{82pt} the power of microfinancing.}\\
    \hline
    \multicolumn{2}{c|}{Top-3 of attention score $\bm{\alpha}^{(2)}\!\!\!\!\!$} & \multicolumn{2}{c|}{Top-3 of attention score $\bm{\alpha}^{(1)}$} & \multicolumn{2}{c}{Top-3 of attention score $\bm{\alpha}^{(0)}$} \\
    \hline
    \textit{/business}: & 0.422 & \textit{NA} & 0.383 & \textit{NA} & 0.387 \\
    \textit{NA}: & 0.384 & \textit{/business/company}: & 0.272 & \textit{/business/company/founders}: & 0.197 \\
    \textit{/location}: & 0.037 & \textit{/business/person}: & 0.063 & \textit{/business/person/company}: & 0.063 \\
    \midrule
    \multicolumn{6}{l}{\textbf{Example Sentence 2:} On sunday, though, there was a significant shift of the tectonic plates of bangladeshi politics, as } \\
    \multicolumn{6}{l}{\hspace{83pt} \textit{muhammad\_yunus}, the founder of a microfinance empire, known as the \textit{grameen\_bank} and the } \\
    \multicolumn{6}{l}{\hspace{83pt} winner of the 2006 nobel peace prize, announced that he would start a new party and step into  } \\
    \multicolumn{6}{l}{\hspace{83pt} the electoral fray.} \\
    % \multicolumn{6}{l}{} \\
    \hline
    \multicolumn{2}{c|}{Top-3 of attention score $\bm{\alpha}^{(2)}\!\!\!\!\!$} & \multicolumn{2}{c|}{Top-3 of attention score $\bm{\alpha}^{(1)}$} & \multicolumn{2}{c}{Top-3 of attention score $\bm{\alpha}^{(0)}$} \\
    \hline
    \textit{/business}: & 0.755 & \textit{/business/company}: & 0.679 & \textit{/business/company/founders}: & 0.652 \\
    \textit{NA}: & 0.103 & \textit{NA}: & 0.089 & \textit{NA}: & 0.069 \\
    \textit{/people}: & 0.031 & \textit{/business/person}: & 0.059 & \textit{/business/person/company}: & 0.057 \\
    \bottomrule
    \end{tabular}
    \caption{Two example sentences with top-3 sent2rel attention scores at all relation levels. Both sentences express the same long-tail relation ``\textit{/business/company/founders}''.}
    \label{tab:casestudy}
\end{table*}

\paragraph{Performance based solely on Sent2rel Module. }
Multi-granular relation labels are used as supervision signals for sent2rel attention modules, and the accuracy of each module is greater than $90\%$ as in Figure \ref{fig:pr_curve}.
Therefore, it is interesting to check if the attention scores can be directly used to predict relations at bag level. We present two settings: 1) only using attention scores on fine-grained relations, i.e., $\bm{\alpha}^{(0)}$, and 2) using products of attention scores at all three levels to make the best of relation hierarchies. As a result, setting 1 and 2 deliver AUC of 0.41 and 0.43 respectively, which surprisingly outperform several previous works in Table \ref{tab:eval_on_benchmark}. 

\paragraph{Error Analysis. }
To investigate the possible reasons for misclassification, we manually check several randomly-sampled error examples from the test set and find the following factors can cause wrong predictions. 
1) Most of error cases demonstrate the proposed model still struggles in handling wrong labeling problem, possibly because limited expressive power of text representation is incompetent at handling noisy, imbalance data. 
2) The sent2rel attention could be invalid when sibling relations have totally distinct meanings, and posts negative effects on relation extraction. For example, \textit{/people/person/children} and \textit{/people/person/profession} refer to opposite meanings.
3) Since a sentence embedding is augmented by multiple semantically-related relation embeddings, relation ambiguity problem deteriorates to post errors. For example, it is hard to distinguish \textit{/people/deceased\_person/place\_of\_death} and \textit{/people/deceased\_person/place\_of\_burial}.

\section{Related Work}
\paragraph{Relation Extraction.} 
% Many works \cite{liu2016learning,du2018multi} are based on selective attention \cite{lin2016neural} framework to handle wrong labeling problem in distantly supervised relation extraction. 
% For example, Ye and Ling~\shortcite{ye2019distant} propose bag-level selective attention to share training information among the bags with the same relation label. Li et al.~\shortcite{li2020self} propose to replace the attention with a gate mechanism especially for one-sentence bags. 
Supervised relation extraction models \cite{zelenko2003kernel,guodong2005exploring} require large amounts of annotated data, which is time consuming and labor intensive. To obtain a large amount of labeled data, Mintz et al.~\shortcite{mintz2009distant} propose distant supervision method to automatically annotate data. However, it inevitably leads to the wrong labeling problem due to the strong assumption. To reduce the effect of wrong labeling problem, multi-instance learning paradigm \cite{riedel2010modeling,hoffmann2011knowledge} is proposed. To introduce the merits of deep learning into relation extraction, 
Zeng et al.~\shortcite{zeng2014relation,zeng2015distant} specifically design the position embedding and piecewise convolutional neural network to better extract the features of each sentence. To further alleviate the effect of wrong labeling problem, Lin et al.~\shortcite{lin2016neural} propose the selective attention framework under multi-instance learning paradigm. Recently, many works \cite{liu2016learning,du2018multi,li2020self} are built upon the selective attention \cite{lin2016neural} framework to handle wrong labeling problem in distant supervision relation extraction. For example, Ye and Ling~\shortcite{ye2019distant} propose bag-level selective attention to share training information among the bags with the same label. 
Hu et al.~\shortcite{hu2019improving} propose a multi-layer attention-based model with joint label embedding.
Li et al.~\shortcite{li2020self} propose to replace the attention with a gate mechanism especially for one-sentence bags.

\paragraph{Hierarchical Relation Extraction.}
More related to our work, to alleviate long-tail problem posted by distant supervision, it is natural to utilize relation hierarchies for knowledge transfer crossing relations. There are two existing works falling into this paradigm. 
Besides using the embedding of fine-grained relation as a query of selective attention, Han et al.~\shortcite{han2018hierarchical} also use embeddings of coarse-grained relations as extra queries to perform a hierarchical attention. 
Zhang et al.~\shortcite{zhang2019long} enhance embeddings of multi-granular relations by merging the embeddings from pre-trained graph model and GCN to alleviate long-tail problem.

\section{Conclusion}  % and Future Work
In this paper, we propose a novel multi-instance learning framework, relation-augmented attention, as our base model for distantly supervised relation extraction. 
It operates at bag level to minimize the effect of the wrong labeling and leverages a sent2rel attention to alleviate the long-tail problem. 
By fully exploiting hierarchical knowledge of relations, we extend the base model to CoRA for boosting the performance on long-tail relations. 
The experiments conducted on the NYT dataset show that CoRA delivers new state-of-the-art performance in terms of both standard metrics (i.e., top-n precision, PR curve, AUC) and long-tail metrics (i.e., Hits@K). And extensive analyses provide comprehensive insights into the proposed model, and verify its capability in handling both wrong labeling and long-tail problems.

\section*{Acknowledgement}
This research was funded by the Australian Government through the Australian Research Council (ARC) under the grant of LP180100654. The authors would like to appreciate anonymous reviewers for their insightful and constructive feedback.

\bibliographystyle{coling}
\bibliography{reference}

\end{document}